\documentclass{article}

\usepackage{PRIMEarxiv}

\usepackage[utf8]{inputenc} 
\usepackage[T1]{fontenc}    
\usepackage[colorlinks=true,citecolor=blue]{hyperref}       
\usepackage{url}            
\usepackage{booktabs}       
\usepackage{amsfonts}       
\usepackage{nicefrac}       
\usepackage{microtype}      
\usepackage{lipsum}
\usepackage{fancyhdr}       
\usepackage{graphicx}       
\usepackage{amsmath}
\usepackage{algorithm}
\usepackage{algorithmic}
\usepackage{multirow}
\usepackage{tabularx}
\usepackage{float}
\usepackage{booktabs}
\usepackage{threeparttable}
\usepackage{xcolor}
\graphicspath{{media/}}     

\ifx\assump\undefined
\newtheorem{assump}{Assumption}
\fi
\title{
Adaptive Liquidity Provision in Uniswap V3 with Deep Reinforcement Learning
}

\author{
  Haochen Zhang\\
  Department of Electrical and Computer Engineering\\
  University of California, Los Angeles \\
  \texttt{haochen235@ucla.edu} \\
  \And
 Xi Chen \\
 Leonard N. Stern School of Business\\
  New York University \\
  \texttt{xc13@stern.nyu.edu} \\
    \And
  Lin F. Yang \\
  Department of Electrical and Computer Engineering\\
  University of California, Los Angeles \\
  \texttt{linyang@ee.ucla.edu} \\
}

\begin{document}
\maketitle

\begin{abstract}
Decentralized exchanges (DEXs) are a cornerstone of decentralized finance (DeFi), allowing users to trade cryptocurrencies without the need for third-party authorization. Investors are incentivized to deposit assets into liquidity pools, against which users can trade  directly, while paying fees to liquidity providers (LPs). 
However, a number of unresolved issues related to capital efficiency and market risk hinder DeFi's further development. 
Uniswap V3, a leading and groundbreaking DEX project, addresses capital efficiency by enabling LPs to concentrate their liquidity within specific price ranges for deposited assets. Nevertheless, this approach exacerbates market risk, as LPs earn trading fees only when asset prices are within these predetermined brackets. 
To mitigate this issue, this paper introduces a deep reinforcement learning (DRL) solution designed to adaptively adjust these price ranges, maximizing profits and mitigating market risks. Our approach also neutralizes price-change risks by hedging the liquidity position through a rebalancing portfolio in a centralized futures exchange. The DRL policy aims to optimize trading fees earned by LPs against associated costs, such as gas fees and hedging expenses, which is referred to as loss-versus-rebalancing (LVR). Using simulations with a profit-and-loss (PnL) benchmark, our method demonstrates superior performance in ETH/USDC and ETH/USDT pools compared to existing baselines. We believe that this strategy not only offers investors a valuable asset management tool but also introduces a new incentive mechanism for DEX designers.

\end{abstract}


\section{Introduction}

Traditionally, intermediaries play essential roles in connecting market participants in transactions. However, with the emergence of decentralized finance (DeFi), decentralized exchange (DEX) - as the most important part of DeFi - enables transactions on a peer-to-peer permissionless network and eliminates the necessity of relying on intermediaries, such as brokers and banks \cite{cartea2022decentralised}. DEX employs an automated market maker (AMM) where the rules to determine exchange rates, as well as to clear demand and supply, are very different from traditional limit order books (LOBs). The constant function market maker (CFMM) is a representative of AMM, and the majority of AMMs belong to this category. CFMMs use a deterministic function $f(x,y)=k$, where $x$ and $y$ represent the reserves of two tokens $X$ and $Y$, and $k$ represents the depth of the pool, and a few additional rules to build an environment that allows traders to swap tokens and also enable liquidity providers (LPs) to deposit tokens. CFMMs with concentrated liquidity (CL) are the most popular type of CFMMs, and they also present LPs with both opportunities and challenges. In CFMMs without CL, LPs provide liquidity uniformly across the entire price range $(0, \infty)$. However, in CFMMs with CL, LPs specify a range of contract prices, i.e., prices of token $X$ with respect to token $Y$ in the Uniswap V3 pool, to provide liquidity by depositing the corresponding assets. When the contract price is within this range, LPs are rewarded with trading fees from traders as a fixed percentage of the transaction size. As a prime example of CFMMs with CL, Uniswap V3 has gained significant attention and has attracted more than $\$9$ billion dollars of value in its liquidity pool. Therefore, a number of papers have studied the approaches on liquidity provision in Uniswap V3\cite{neuder2021strategic,bar2023uniswap,fritsch2021concentrated,fan2022differential,heimbach2022risks,cartea2022decentralised}.

\begin{table}[H]
    \centering
    
    \caption{Different objective functions in designing liquidity provision strategies.}
    \label{tab:measure of PnL}
    
    \begin{tabular}{cc}
    \toprule
    Reference & Objective Function\\
    \midrule
    
    \cite{neuder2021strategic} & trading fee\\
    \cite{bar2023uniswap} & trading fee + change of position value\\
    \cite{fritsch2021concentrated} & trading fee + change of position value\\
    \cite{fan2022differential} & trading fee + impermanent loss\\
    \cite{heimbach2022risks} & trading fee + impermanent loss \\
    \cite{cartea2022decentralised} & trading fee + loss-versus-rebalancing\\
    \textbf{this study} & trading fee + loss-versus-rebalancing + gas fee \\
    \bottomrule
    
    \end{tabular}

\end{table}
To optimally manage liquidity provision, it is essential to construct an objective function that quantifies the return on investment (ROI) for a liquidity provider (LP). This objective function must account for all relevant costs and returns. For the purpose of this discussion, we assume that both the initial investment and the final ROI are denominated in U.S. dollars (USD).
One primary cost incurred by an LP is the blockchain gas fee, significant during the reallocation of liquidity. These fees should not be underestimated, as they substantially influence the frequency of liquidity reallocation. Another variable cost involves changes in position value, attributed to asset price fluctuations.
Despite these considerations, existing literature often falls short in offering an accurate profit and loss (PnL) definition for an LP. Table~\ref{tab:measure of PnL} highlights the objective functions adopted in existing works. Notably, many studies, such as \cite{neuder2021strategic}, neglect gas fees, while others incorporate metrics that do not capture an LP's true PnL comprehensively.

To rectify this, we propose a PnL definition that includes trading fees, gas fees, and changes in position value, aligning closely with the LP's actual ROI. However, the absence of an effective hedging mechanism can make the returns unstable due to market volatility.
To mitigate this, we propose to adopt a hedging strategy by taking an opposing position in a separate market, such as a futures market. In this framework, we introduce a new PnL metric called Loss-Versus-Rebalancing (LVR), which is defined as the value difference between a rebalanced portfolio and the original liquidity position \cite{milionis2022automated}. Combined with the gas fee and trading fee, our strategy ensures a more realistic ROI despite market fluctuations.

In sum, this paper sets out to accomplish two primary goals:
\begin{enumerate}
    \item To hedge liquidity provision risks effectively through a rebalancing portfolio, thereby addressing the risk of liquidity provision due to the change of position value and only leaving a residual called LVR which can be seen as the cost of providing liquidity \cite{milionis2022automated,cartea2022decentralised}.
    \item To employ Deep Reinforcement Learning (DRL) algorithms for decision-making. DRL is favored as it is model-free and offers greater flexibility compared to traditional parameterized models like geometric Brownian motion \cite{bar2023uniswap,fritsch2021concentrated}.
\end{enumerate}

The main contributions of this paper can be summarized as follows:

\textbf{(1)} We present a more pragmatic measure of PnL tailored specifically for LPs in Uniswap V3. This comprehensive metric includes trading fees, LVR, and gas fees. When LPs hedge their liquidity positions through a rebalancing portfolio, this metric precisely reflects their actual profits.

\textbf{(2)} Our strategy is pioneering in its systematic exploration of a DRL-based strategy for liquidity provision in Uniswap V3. We empirically demonstrate the effectiveness of DRL for optimizing liquidity provision, particularly when market risks are hedged. Our simulation results corroborate that optimizing the PnL of an unhedged liquidity position presents a more complex challenge compared to a hedged position.

\textbf{(3)} Our proposed method enhances liquidity provision strategies, thereby increasing the potential profits for LPs. Importantly, this strategy does not require substantial computational resources and is thus accessible to individual investors. Furthermore, our approach shows considerable advantages over baseline methods, especially for investors with limited initial capital.

The structure of this paper is organized as follows: Section~\ref{sec:related work} reviews a broad range of applications of RL in finance.
 Section~\ref{sec:preliminary} delineates the Uniswap V3 contract mechanics, introduces the concept of LVR, and provides an overview of RL and the specific DRL algorithm employed in this study.
Section~\ref{sec:method} outlines how the liquidity provision problem in Uniswap V3 is formulated as a reinforcement learning problem.
Section~\ref{sec:experiment} discusses the experimental setup and presents key findings from simulations conducted on historical Uniswap V3 data.
Lastly, Section~\ref{sec:discussion and conclusion} concludes.

\begin{figure}[!t]
  \centering
  \includegraphics{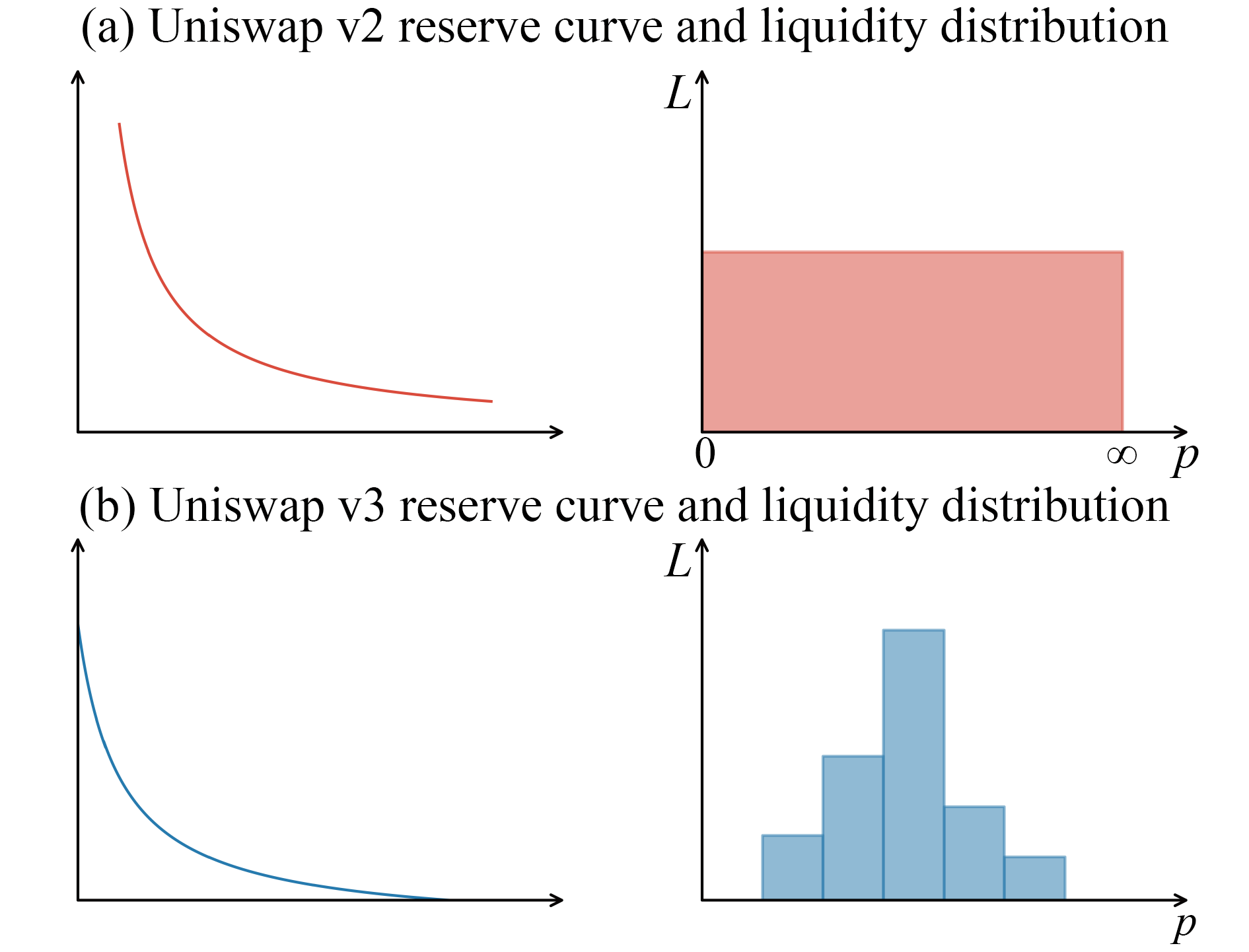}
  \caption{A toy example to show the difference in reserve curve and liquidity distribution between Uniswap V2 and Uniswap V3}
  \label{reserve curve}
\end{figure}

\section{Related Work \label{sec:related work}}

This section provides an overview of the existing literature on the application of DRL in finance, with a focus on domains other than liquidity provision in Uniswap V3. The literature can be broadly categorized into four key areas: Optimal Execution, Portfolio Optimization, Option Pricing and Hedging, and AMMs.

Optimal execution involves the strategic buying or selling of a given asset within a specific time frame while minimizing market impact and price uncertainty. A notable contribution in this area is by \cite{moallemi2022reinforcement}, which employs both supervised learning and reinforcement learning techniques to develop an optimal stopping formulation for this problem.

Portfolio Optimization addresses the challenge of maximizing expected returns while controlling the risk across a basket of assets. \cite{park2020intelligent} employs a deep Q-learning agent that outperforms traditional benchmarks. It effectively tackles the dimensionality issue by using a discrete action space and introduces a mapping function to handle infeasible actions. \cite{liu2021finrl} streamlines automated trading strategies with their \textit{FinRL} framework.

In options trading, the goal is often to balance the costs and benefits of various hedging strategies. While classical models like Black-Scholes \cite{black1973pricing} and Merton \cite{merton1973theory} offer theoretical insights, these models often fall short in real-world conditions due to trading frictions. \cite{kolm2019dynamic} explores this challenge and demonstrates that reinforcement learning can effectively optimize hedging strategies under realistic market conditions.

In the rapidly evolving field of AMMs, reinforcement learning is also gaining traction for designing more efficient market-making algorithms. For instance, \cite{sabate2022case} employs an actor-critic method to explore the profitability of liquidity provision in constant product markets, suggesting the inclusion of flexible fee tiers for liquidity providers. Complementing this, \cite{lim2022predictive} integrates LSTM with Q-learning to improve the performance of predictive AMMs, building upon the foundation laid by Uniswap V3.

\section{Preliminary \label{sec:preliminary}}

\subsection{Uniswap V3}

A pool in Uniswap V3 is specified by three features: token1, token2, and fee tier. Uniswap V3 discretizes prices using ticks, where the price corresponding to the $i^\mathrm{{th}}$ tick is $1.0001^{i}$. Tick spacing, the finest width of the liquidity interval, is determined by the fee tier \cite{elsts2021liquidity}. For instance, 1\% pools have a tick spacing of 200 ticks, 0.3\% pools have a tick spacing of 60 ticks, and 0.05\% pools have a tick spacing of 10 ticks. This paper focuses on Uniswap V3 pools with a stable coin and a volatile coin, but can be expanded to pools with two volatile coins without difficulty. We assume token1 is a volatile coin and token2 is a stable coin herein. We use \textit{X}, \textit{Y}, $x$, $y$, $\delta$, and $d$ to represent the volatile coin, the stable coin, reserve of the volatile coin, reserve of the stable coin, fee tier, and tick spacing, respectively. 

The key difference between Uniswap V3 and Uniswap V2 lies in the introduction of Concentrated Liquidity (CL) in Uniswap V3, leading to a non-uniform distribution of liquidity, as illustrated in Figure~\ref{reserve curve}. In Uniswap V2, the reserve curve is straightforward and neat, described by the equation $x\cdot y=L^2$. In contrast, the reserve curve in Uniswap V3 operates within a defined price range $\left[ p_a, p_b\right]$, where a consistent amount of liquidity units denoted as \textit{L} is maintained. The relationship between the reserves of assets \textit{X} and \textit{Y} is expressed as $\left(x+L/\sqrt{p_b}\right)\cdot \left(y+L\sqrt{p_a}\right) = L^{2}$. The contract price of asset \textit{X}, represented as $p$, can be calculated using the equation $p=\left(y+L\sqrt{p_a}\right)/\left(x+L/\sqrt{p_b}\right)$ when the price falls within the interval $\left[ p_a, p_b\right]$. The reserves of assets \textit{X} and \textit{Y} within a pool can be expressed in terms of \textit{p}, as shown in equation (\ref{eq1}).

\begin{equation}
\label{eq1}
   \left( x, y  \right)=
    \begin{cases}
      \left ( L\left(\frac{1}{\sqrt{p_a} }-\frac{1}{\sqrt{p_b} } \right),0 \right ) & p \le p_a\\
      \left ( L\left(\frac{1}{\sqrt{p} }-\frac{1}{\sqrt{p_b} } \right),L\left( \sqrt{p}-\sqrt{p_a} \right )  \right ) & p_a < p < p_b\\
      \left ( 0,L\left ( \sqrt{p_b}-\sqrt{p_a} \right )  \right ) & p\ge p_b
    \end{cases}       
\end{equation}
Then we consider how to calculate trading fees for an LP who has a liquidity position with \textit{L} units of liquidity in the range $\left[p_a,p_b\right]$. If a swap happens and the contract price moves from $p$ to $p^{\prime}$. The LP can obtain a trading fee from the trader who initiates the swap in two scenarios as depicted in equation (\ref{condition1}). The first case corresponds to an upward price movement when a trader holds \textit{Y} and swaps it for \textit{X}, whereas the second case involves a downward price movement when some \textit{X} flows into the pool and some \textit{Y} flows out. 
\begin{equation}
\label{condition1}
\begin{cases}
    \left[p_a,p_b\right] \cap \left[p,p^{\prime}\right] \ne \phi & \mathrm{if \ } p \le p^{\prime}\\
    \left[p_a,p_b\right] \cap \left[p^{\prime},p\right] \ne \phi & \mathrm{if \ } p > p{\prime} \\    
\end{cases}   
\end{equation} 
\par 

To better explain how to calculate trading fees, let's consider a specific scenario where both \textit{p} and $p^{\prime}$ fall within the interval $\left[p_a,p_b\right]$, and the liquidity distribution in this interval is uniform, serving as a simple example. If $p \le p^{\prime}$, then the change in the reserve of $Y$ in the liquidity pool, denoted by $\Delta y$, is positive and can be calculated using (\ref{eq1}): $\Delta y=L_{\mathrm{total}}\left(\sqrt{p^{\prime}}-\sqrt{p}\right)$. Here, $L_{\mathrm{total}}$ represents the total liquidity within $\left[p_a,p_b\right]$. Since a proportion of $(1-\delta)$ of $Y$ enters the liquidity pool, while a proportion of $\delta$ of $Y$ is skimmed as LPs' fee, the total fee for all LPs holding liquidity in the range $\left[p_a,p_b\right]$ is $\frac{\delta \Delta y}{1-\delta}=\frac{\delta}{1-\delta} L_{\mathrm{total}}\left(\sqrt{p^{\prime}}-\sqrt{p}\right)$. The total trading fee is distributed to LPs in proportion to the amount of liquidity units they provide. Thus, the trading fee for an LP holding $L$ liquidity units is $\frac{\delta}{1-\delta} L\left(\sqrt{p^{\prime}}-\sqrt{p}\right)$, and this value is independent of $L_{\mathrm{total}}$. In the event of a downward price movement, the trading fee value in terms of $Y$ for the LP can be calculated in a similar manner \cite{fan2022differential}. Equation (\ref{eq2}) summarizes the formulas for the two cases mentioned above.


\begin{equation}
\label{eq2}
   fee=
    \begin{cases}
    \frac{\delta}{1-\delta} L\left(\sqrt{p^{\prime}}-\sqrt{p}\right) & \mathrm{if \ } p \le p^{\prime}\\
    \frac{\delta}{1-\delta} L\left(\frac{1}{\sqrt{p^{\prime}}}-\frac{1}{\sqrt{p}}\right)p^{\prime} & \mathrm{if \ } p > p{\prime} \\
    \end{cases}       
\end{equation}
\par 

\begin{figure}[ht]
  \centering
  \includegraphics{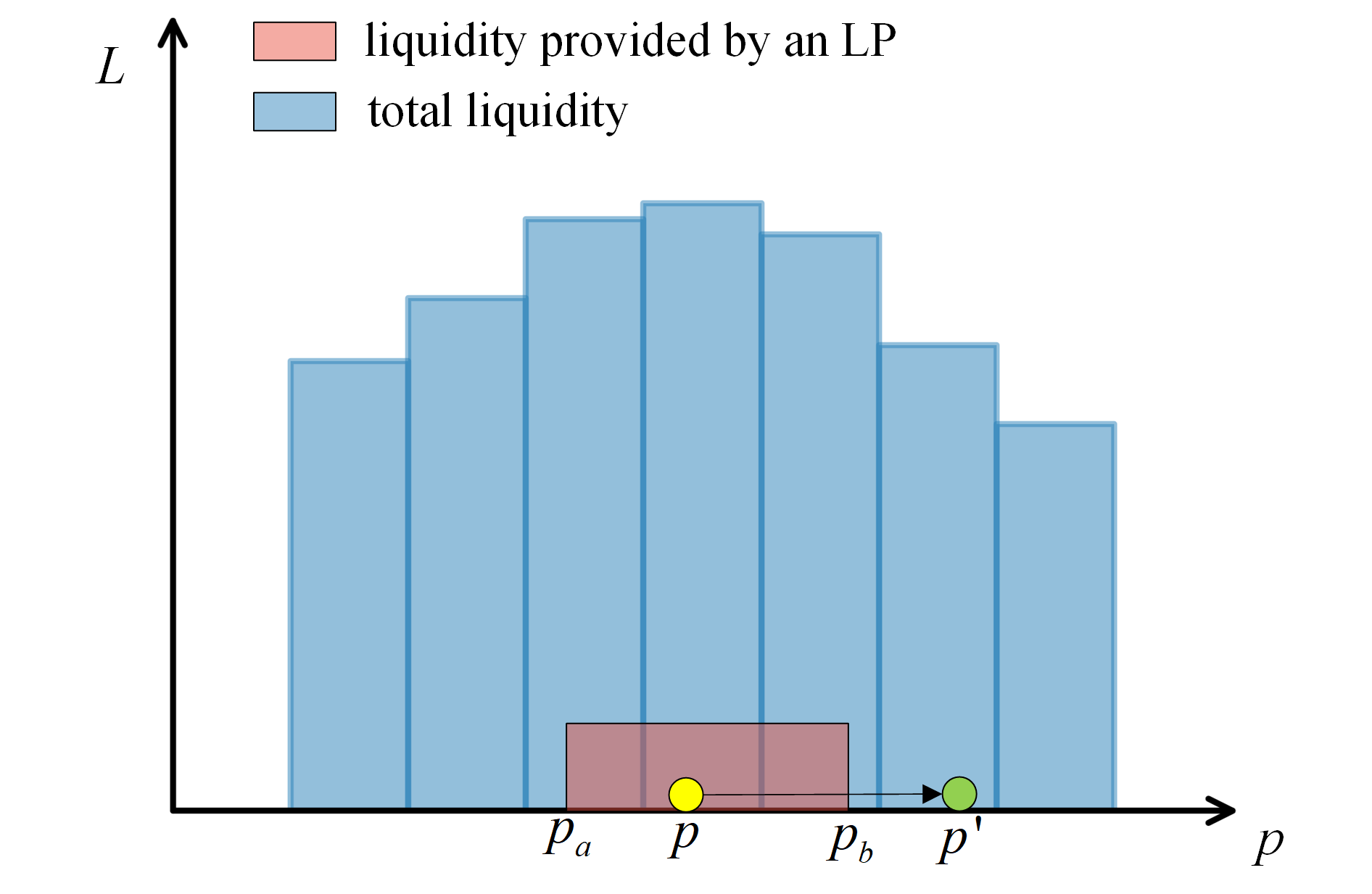}
  \caption{An example of swap decomposition when calculating LP's income}
  \label{decompose swap}
\end{figure}

In a broader context, it is essential to acknowledge that (\ref{condition1}) may not hold true. However, a swap can be decomposed into several parts if the price movement extends beyond the bounds of an LP's liquidity position \cite{fan2022differential}. As shown in Figure~\ref{decompose swap}, a price movement from $p$ to $p^{\prime}$ when $p\in \left[p_a,p_b\right]$ but $p^{\prime}\notin \left[p_a,p_b\right]$ can be decomposed into two sequential movements: first from $p$ to $p_b$, and then from $p_b$ to $p^{\prime}$. And the LP is rewarded with trading fees only in the first price movement from $p$ to $p_b$. It is noteworthy that a price movement leads to a change in the liquidity position's value, which can be expressed as equation~(\ref{eq4}), where $x^{\prime}$ and $y^{\prime}$ can be calculated by substituting $p^{\prime}$ into equation~(\ref{eq1}).

\begin{equation}
\label{eq4}
    \Delta V = p^{\prime} x^{\prime}  + y^{\prime} - \left(px + y\right)
\end{equation}

\subsection{Loss-Versus-Rebalancing of A Liquidity Provider \label{sec:lvr_and_pnl}}

Loss-versus-rebalancing (LVR) is proposed by \cite{milionis2022automated}, where the PnL of a liquidity position without receiving any trading fee is shown to be composed of the PnL of a rebalancing strategy that can be hedged by its opposite position and a predictable residue that represents the cost of providing liquidity, i.e., LVR. For clarity, this subsection first defines LVR for a discrete price trajectory, which directly determines the reward function of a DRL agent, then recapitulates the derivation of LVR under the setting of a geometric Brownian motion contract price and explains LVR is always non-positive.

Given a discrete trajectory of contract price $\left \{ p_t \right \}_{t=1...N} $, LVR can be expressed by (\ref{eq7}).
\begin{equation}
\label{eq7}
\begin{split}
\mathrm{LVR} = &\sum_{t=1}^{N-1} \{ V\left(p_{t+1}\right)-V\left(p_t\right)- x\left(p_t\right)\left(p_{t+1}-p_t\right) \}\\
= &\sum_{t=1}^{N-1} \{p_{t+1}\left[x\left(p_{t+1}\right)-x\left(p_{t}\right)\right]+y\left(p_{t+1}\right)-y\left(p_{t}\right)\}
\end{split}
\end{equation}

where $V(p_{t+1}) - V(p_t)$ represents the change of position value, and $-x(p_t)(p_{t+1}-p_t)$ is the return of shorting rebalancing portfolio.

Based on equation~(\ref{eq7}), to avoid the market risk of a liquidity position, one can invest in a portfolio that consists of a fixed liquidity position and the opposite position of its corresponding rebalancing strategy. The liquidity position is used to earn trading fees in Uniswap V3, whereas the short position is for hedging the market risk of the liquidity position. Therefore, the PnL of the whole portfolio can be defined as $\mathrm{PnL_{static}}=\mathrm{Fee}+\mathrm{LVR}$, where Fee is the cumulative trading fee during the period when the liquidity position is not adjusted from $t=1$ to $t=N$.


For a given liquidity position, i.e., with fixed $p_a$, $p_b$, and $L$, its value can be written as equation (\ref{v(pt)}), where $x(p_t)$ and $y(p_t)$ denote the amounts of $X$ and $Y$ reserved in the liquidity position. If the contract price evolves according to a geometric Brownian motion, i.e., $dP_{t}/P_{t}=\sigma dB_t$, where $B_t$ is a Brownian motion, $dV_t$ can be obtained by applying the It$\mathrm{\hat{o}}$ lemma to $V\left(p_t\right)$ \cite{milionis2022automated}, as shown in equation~(\ref{eq5}).

\begin{equation}
\label{v(pt)}
   V(p_t)=p_t x(p_t)+y(p_t)=
    \begin{cases}
      L \left(\frac{1}{\sqrt{p_a} }-\frac{1}{\sqrt{p_b} } \right) p_t & p \le p_a\\
      2 L \sqrt{p_t} - \frac{L}{\sqrt{p_b}} p_t - L \sqrt{p_a}  & p_a < p < p_b\\
      L\left ( \sqrt{p_b}-\sqrt{p_a} \right ) & p\ge p_b
    \end{cases}       
\end{equation}

\begin{equation}
\label{eq5}
\begin{split}
         dV(p_t) &= V^{\prime}\left(p_t\right)dp_t+V^{\prime\prime}\left(p_t\right)\left(dp_t\right)^2
     \\ &= V^{\prime}\left(p_t\right)dp_t+V^{\prime\prime}\left(p_t\right)\sigma^2p_t^2dt
     \\ &= x(p_t)dp_t+V^{\prime\prime}\left(p_t\right)\sigma^2p_t^2dt
\end{split}
\end{equation}

where $V^{\prime}\left(p_t\right)$ and $V^{\prime \prime}\left(p_t\right)$ denote the first derivative and the second derivative of $V$ with respect to $p_t$.

A rebalancing strategy with respect to a liquidity position is a self-financing strategy whose holdings are defined by a stochastic process $\left ( x_{r}(t), y_{r}(t) \right )$, representing the holdings of $X$ and $Y$ in the rebalancing portfolio at time $t$, where $x_{r}(t)=x(p_t)$ for all $t$. Under the assumption of frictionless trading and in the absence of trading fees, the PnL of a rebalancing strategy is $x(p_t)dp_t$, precisely mirroring the first term on the right-hand side of (\ref{eq5}). Therefore, a part of $dV\left(p_t\right)$ can be hedged by taking the opposite position of the rebalancing portfolio\cite{milionis2022automated}. By taking the second derivative of equation (\ref{v(pt)}), $V^{\prime\prime}\left(p_t\right)$ can be expressed by equation (\ref{eq6}); and it is always non-positive. Hence the second term on the right-hand side of (\ref{eq5}) can be regarded as a loss term and it is defined as loss-versus-rebalancing (LVR). Therefore, LVR can be regarded as the expense associated with providing liquidity, while the PnL of the rebalancing strategy can be seen as the market risk of providing liquidity.
\begin{equation}
\label{eq6}
   V^{\prime\prime}\left(p_t\right)=
    \begin{cases}
     -\frac{L}{2p_t^{1.5}}& p_a < p_t < p_b\\
     0 & \mathrm{otherwise} \\
    \end{cases}       
\end{equation}

\subsection{Deep Reinforcement Learning}
Given that the liquidity provision problem is structured as a Markov decision process with a discrete action space, this research employs the Dueling DDQN \cite{wang2016dueling}, to acquire adaptive strategies for liquidity provision. In this section, we commence by introducing the notations commonly utilized in general reinforcement learning problems. Subsequently, we delve into the fundamental principles of Dueling DDQN.

We formulate the optimal liquidity provision problem as a policy search problem in a Markov decision process denoted by $\left(\mathcal{S}, \mathcal{A}, \mathcal{T}, r\right)$. $\mathcal{S}$ and $\mathcal{A}$ denote state space and action space, $\mathcal{T}:\mathcal{S}\times \mathcal{S} \times \mathcal{A} \times \mathbb{R} \to \mathbb{R}$ represents state transition probability, i.e., dynamic of the environment, $\mathcal{T}\left(s^{\prime},r,s,a \right)=Pr\{S_{t+1}=s^{\prime}, R_t=r| S_t=s, a_t=a\}$. $r:\mathcal{S} \times \mathcal{A} \to \mathcal{R}$ denotes reward function, $r(s_t,a_t)$ is reward received by the agent after make action $a_t$ at state $s_t$. An agent's action is determined by a policy $\pi: \mathcal{S} \times \mathcal{A} \to \left[0,1\right]$, where $\pi$ must satisfy $\sum_{a_t\in \mathcal{A}}\pi\left(a_t|s_t\right)=1$. $\pi\left(a_t|s_t\right)$ represents the probability of taking action $a_t$ at state $s_t$. The optimal policy maximizes discounted return defined as $G_t=\sum_{\tau=t+1}^{\infty}\gamma^{\tau-t-1}r_{\tau}$, where $\gamma \in (0,1)$ is discount factor. State-value function is defined as $ V^{\pi}(s) = \mathbb{E}_{\pi,\mathcal{T}}\left[G_t|S_t=s\right]$ and action-value function is defined as $Q^{\pi}(s,a)=\mathbb{E}_{\pi,\mathcal{T}}\left[G_t|S_t=s,A_t=a\right]$. The optimal action-value function is defined as 
\begin{displaymath}
    Q^{\ast}(s,a)={\max_{\pi} } \mathbb{E}_{\pi,\mathcal{T}}\left[G_t|S_t=s,A_t=a\right]
\end{displaymath}
$Q^{\ast}$ satisfies Bellman optimality equation, as shown in (\ref{eq8}). Advantage function is defined as $A^{\pi}(s,a)=Q^{\pi}(s,a)-V^{\pi}(s)$, it provides a relative measure of the importance of each action. One way to estimate $Q^{\ast}(s,a)$ is using the Bellman optimality equation to do an iterative update, as shown in (\ref{eq9}) \cite{sutton2018reinforcement}.
\begin{equation}
\label{eq8}
\begin{split}
Q^{\ast}(s,a)=\sum_{s^{\prime},r}\mathcal{T}\left(s^{\prime},r,s,a \right)\left[r+\gamma {\max_{a^{\prime}}}Q^{\ast}(s^{\prime},a^{\prime})\right]
\end{split}
\end{equation}

\begin{equation}
    \label{eq9}
    Q^{(i+1)}(s,a)\gets \sum_{s^{\prime},r}\mathcal{T}\left(s^{\prime},r,s,a \right)\left[r+\gamma {\max_{a^{\prime}}}Q^{(i)}(s^{\prime},a^{\prime})\right]
\end{equation}

Deep Q-learning is a value-based algorithm that uses a neural network (DQN) to estimate action-value function, i.e., $\hat{Q}(s,a;\theta)\approx Q^{\ast}(s,a)$. In particular, Deep Q-learning collects transitions by applying $\epsilon$-greedy policy and stores them in a replay buffer, and after making an action at a state $s_t$, it samples a minimatch of transitions from the replay buffer and performs gradient descent to minimize loss function defined by (\ref{eq10}). Details of DQN can be seen in \textbf{Algorithm 1} in \cite{mnih2013playing}. 
\begin{equation}
\label{eq10}
    L(\theta)=\mathbb{E}_{(S_t, A_t)\sim D}\left[\left(\hat{Q}(S_t,A_t;\theta)-Y_t^\mathrm{DQN}\right)^{2}\right]
\end{equation}
where $Y_t^{\mathrm{DQN}}=R_t+\gamma {\max_{a^{\prime}}}\hat{Q}(S_{t+1},a^{\prime};\theta)$ is the target for a transition from the replay buffer.

\cite{hasselt2010double} shows that using the same value for selection and evaluation leads to overoptimistic estimates. And \cite{van2016deep} shows that overestimations in DQN is severe and proposes double DQN borrowing the idea from double Q-learning. Double DQN adopts a target network and replace the target with $Y^{\mathrm{Q}}=R_t+\gamma \hat{Q}
(S_{t+1},\mathrm{argmax}_{a}\hat{Q}(S_{t+1},a;\theta^{\prime});\theta)$, where $\theta$ and $\theta^{\prime}$ are parameters of local Q-network and target Q-network, respectively. Dueling DDQN \cite{wang2016dueling} is a further improvement of DDQN that introduces a dueling network architecture, which estimates action-value function by applying (\ref{eq11}). Pseudo-code of Dueling DDQN can be seen in \textbf{Algorithm 1} in \cite{wang2016dueling}.
\begin{equation}
\label{eq11}
\begin{split}
    \hat{Q}\left(s,a;\theta,\alpha,\beta \right) = &\hat{V}\left(s;\theta,\beta \right) + \left[\hat{A}\left(s,a;\theta,\alpha \right)-\frac{1}{\left | \mathcal{A}  \right |}\sum_{a^{\prime}}\hat{A}\left(s,a^{\prime};\theta,\alpha \right)\right]
\end{split}
\end{equation}

\section{Method \label{sec:method}}

\subsection{Optimal Liquidity Provision}


The optimal liquidity provision study focuses on the strategy that maximizes the profit of an LP who holds a certain amount of capital and participates in providing liquidity in Uniswap V3. This problem is studied in discrete time in this context, allowing an LP to adjust liquidity positions at evenly distributed time steps $t \in \{1,2,...,T\}$.

\subsection{Reinforcement Learning Formulation}

This subsection provides a description of states, actions, rewards, and system dynamics in the reinforcement learning formulation of optimal liquidity provision in Uniswap V3. The time period is divided into hourly intervals, and LPs make decisions at the end of each hour. To reduce the learning difficulty for RL agents, this paper does not consider a general liquidity provision strategy that can hold a liquidity position at any tick range, which is intractable due to the dimensionality of state and action. Instead, it focuses on studying the optimal uniform interval reallocation strategy. Under this problem setting, an LP can hold a liquidity position in a tick interval $\left[tick_l, tick_r\right]$ with liquidity units of $L$. At the end of each hour, the LP can choose whether to reallocate the liquidity position.

\begin{table}[!t]
\centering
  \caption{Set of features adopted to construct state}
  \label{tab:features}
  \begin{tabular}{p{8cm}c}
    \toprule
    Feature Name & number of features\\
    \midrule
hourly open price &  1  \\
hourly highest price/hourly open price &  1  \\
hourly lowest price/hourly open price & 1  \\
hourly close price/hourly open price & 1\\
hourly trading volume in USD & 1\\
Double Exponential Moving Average/hourly open price & 1\\
Parabolic SAR/hourly open price & 1\\
Average Directional Movement Index & 1\\
Absolute Price Oscillator & 1\\
Aroon Oscillator & 1\\
Balance Of Power & 1\\
Commodity Channel Index & 2\\
Chande Momentum Oscillator & 1\\
Directional Movement Index & 1\\
Minus Directional Movement&1\\
Momentum&1\\
Plus Directional Movement&1\\
TRIX&1\\
Ultimate Oscillator&1\\
Stochastic Momentum Indicator&3\\
Stochastic Fast Momentum Indicator&3\\
Normalized Average True Range&1\\
True Range&1\\
Hilbert Transform - Dominant Cycle Period &1\\
Hilbert Transform - Dominant Cycle Phase&1\\
\bottomrule
\end{tabular}
\end{table}



\subsubsection{State}
The state at time step $t$, denoted as $s_t \in \mathcal{S}$, consists of market information and the state of the liquidity position. Specifically, we define $s_{t} = \left[ \mathbf{f}_{t}, c_{t}, m_{t}, w_{t}, l_{t} \right] \in \mathbb{R}^{32}$, where $\mathbf{f}_{t} \in \mathbb{R}^{28}$ and $c_{t}$, $m_{t}$, $w_t$, and $l_{t}$ are real numbers. $\mathbf{f}_{t}$ represents the market information features at time step $t$. $c_{t}$ is the amount of USD the agent holds at time step $t$. $m_{t}$ denotes the central tick of the liquidity position, while $w_{t}$ represents the width of the liquidity interval. $l_{t}$ is the value of the liquidity position at time step $t$, measured in USD. In the experiment, $l_0$ is set to different values to demonstrate the differences in optimal liquidity provision with varying initial funds.

The feature vector $\mathbf{f}_{t}$ includes both basic market information, such as hourly open, high, low, and close prices, and technical indicators that incorporate information about the market trend in the past few hours, as presented in Table~\ref{tab:features}. 
Using $w_{t}$ and $m_{t}$, the lower bound and the upper bound of an LP's liquidity position can be determined by $m_t - d \cdot w_t$ and $m_t + d \cdot w_t$. The amount of liquidity is determined by variables such as $w_{t}$, $m_{t}$, $l_t$, and the hourly close price. Therefore, the information integrated into the state $s_t$ is comprehensive.


\subsubsection{Action}
In Uniswap V3, continuous prices are mapped to discrete ticks. The action space in this study is accordingly set to be discrete: $\mathcal{A} = \{0, 1, 2, \ldots, N_a\}$, where $N_a$ represents the maximum width of the interval. An action $a_{t} = 0$ indicates that the agent does not make any adjustment to the liquidity position and maintain the current liquidity position. For $a_t > 0$, the agent intends to withdraw its liquidity investment, reinvesting the funds (including trading fees earned) into a new liquidity position centered at the current tick with a width of $w_{t+1} = a_t$.


\subsubsection{Reward \label{reward_define}}
This study adopts two settings and the corresponding two reward functions. In the first setting, the LP can reallocate the liquidity position and rebalance the short position to hedge against market risk related to the liquidity position. In this context, it is necessary to consider the trading fee and LVR of a fixed liquidity position during the time between the end of the $\mathrm{t^{th}}$ hour and the end of the $\mathrm{t+1^{th}}$ hour, along with the gas fee incurred by the reallocation at the end of the $\mathrm{t^{th}}$ hour. These factors are taken into account when calculating the reward for each transition $\left(s_t, a_t, r_t, s_{t+1}\right)$. Therefore, the reward for a transition is defined as $r_{t} := -\mathbf{1}_{\{a_t\neq 0\}} + \mathrm{Fee}_t + \mathrm{LVR}_t$, where $\mathbf{1}_{\{\cdot\}}$ is the indicator function, and $-\mathbf{1}_{\{a_t\neq 0\}}$ denotes the gas fees incurred by liquidity reallocation. The values of $\mathrm{Fee}_t$ and $\mathrm{LVR}_t$ can be computed using the formulas mentioned in Section~\ref{sec:lvr_and_pnl}.

In the second setting, where an LP reallocates the liquidity position without hedging, it is necessary to consider the trading fee and change in position value of a fixed liquidity position during the time between the end of the $\mathrm{t^{th}}$ hour and the
end of the $\mathrm{t+1^{th}}$ hour, along with the gas fee incurred by the reallocation at the end of the $\mathrm{t^{th}}$ hour. Therefore the reward function is defined as $r_{t} := -\mathbf{1}_{\{a_t\neq 0\}} + \mathrm{Fee}_t + \Delta V_{t}$.

\subsubsection{System Dynamics Modeling}
To let readers better understand reinforcement learning modeling herein, the authors give a simple example that the agent takes action $a_t \neq 0$ to explain the environment dynamics. Given state $s_{t}=\left[ \mathbf{f}_{t}, c_{t}, m_{t}, w_{t}, l_{t}\right]$ and $a_t$, $s_{t+1}=\left[ \mathbf{f}_{t+1}, c_{t+1}, m_{t+1}, w_{t+1}, l_{t+1}\right]$ can be obtained by the following procedure: \\
\textbf{1)}: \ $m_{t+1} \gets d \cdot \mathrm{round} \left ( \mathrm{price2tick}(c_t),d \right )$ \\
\textbf{2)}: \ $w_{t+1} \gets a_t$ \\
\textbf{3)}: \ calculate $u$ using $c_{t}+l_{t}$, $m_{t+1}$, $w_{t+1}$, and $\mathrm{close}_t$ \\
\textbf{4)}: \ obtain $\mathbf{f}_{t+1}$ after moving to the end of the $\mathrm{(t+1)}^{th}$ hour\\
\textbf{5)}: \ $c_{t+1} \gets c_{t} + \mathrm{Fee}_t$ \\
\textbf{6)}: \ calculate $l_{t+1}$ by using $m_{t+1}$, $w_{t+1}$, $u$, and $\mathrm{close}_{t+1}$. \\
where $\mathrm{close}_t$ denotes the close contract price at the $\mathrm{t}^{th}$ hour, $u$ denote the amount of liquidity units, $\mathrm{price2tick}(\cdot)$ denotes the function that converts contract price to tick.

\section{Experiment \label{sec:experiment}}
\subsection{Experiment setting, Baselines, and Indices for Evaluation}

In the experiments presented herein, a few assumptions are adopted as follows:

\begin{assump}
\label{ass:1}
  The gas fee for each liquidity reallocation, which encompasses the sum of the gas fees to withdraw liquidity, adjust holdings, and invest liquidity again, is assumed to be a flat fee. 
\end{assump}

\begin{assump}
\label{ass:2}
  Dynamic trading applying the rebalancing strategy is conducted without friction.
\end{assump}

\begin{assump}
\label{ass:3}
    Slippage is neglected when the agent adjusts holdings after each liquidity withdrawal.
\end{assump}

\begin{assump}
\label{ass:4}
  The actions taken by the agent have no effect on the behaviors of other players in the market.
\end{assump}

Assumption~\ref{ass:1} is reasonable and this assumption is also adopted in \cite{cartea2022decentralised}. Assumption~\ref{ass:2} is reasonable because trading friction on Binance is at most 0.1\% and the trading volume of a rebalancing strategy is small.
Assumption~\ref{ass:3} is reasonable when the LP's position value is not too large.
Assumption~\ref{ass:4} is reasonable when the LP's position value is not too large, and it is necessary for simulation based on historical data.


The proposed method is a Dueling DDQN-based approach that can adaptively control both the hold period and the liquidity spread. During training, the Dueling DDQN method employs an $\epsilon$-greedy strategy, while during testing, it employs a deterministic strategy. The baseline methods include the uniform $\tau$-reset strategy \cite{neuder2021strategic}, the exponential weights adaptive strategy (EWA) \cite{bar2023uniswap}, and the dynamic programming-based method proposed by Cartea et al. \cite{cartea2022decentralised}. The differences between the proposed method and the baselines are outlined in Table~\ref{tab:baselines}. Specifically, the uniform $\tau$-reset strategy involves allocating liquidity uniformly over a range of $\tau$ tick ranges centered on the current contract price and resetting when the contract price moves out of the liquidity interval. Thus, the uniform $\tau$-reset strategy can control the hold period, but the liquidity spread is determined by the hyperparameter $\tau$. EWA adopted herein differs slightly from the original version; it does not reallocate the liquidity interval every hour, but introduces an additional hyperparameter $T_{re}$ which denotes the minimum number of hours the agent holds the liquidity position after each reallocation. EWA is a bandit algorithm that cannot balance short-term and long-term profit. The pseudo-code for EWA is provided in \textbf{Algorithm}~\ref{alg:bandit} in Appendix~\ref{set:hyper}. The dynamic programming-based method from \cite{cartea2022decentralised} can adaptively adjust the liquidity spread but is unable to determine a suitable time to reallocate the liquidity position. For simplicity, the methods are referred to as M1$\sim$M4 in the subsequent content.

\begin{table}[!t]
    \centering
    \caption{Comparison between The Proposed Method and Baselines}
    \label{tab:baselines}
    \begin{tabular}{cccp{2cm}<{\centering}p{2cm}<{\centering}p{2cm}<{\centering}}
    \toprule
    &Methods & Adaptive? & Controllable Hold Period? & Controllable Liquidity Spread? & Consider Gas Fee? \\
    \midrule
    M1 & the proposed method & \checkmark & \checkmark & \checkmark & \checkmark \\
    M2 & uniform $\tau$-reset strategy\cite{neuder2021strategic} & $\times$ & \checkmark & $\times$ & $\times$  \\
    M3 & EWA\cite{bar2023uniswap} & \checkmark & $\times$ & \checkmark & $\times$\\
    M4 & dynamic programming\cite{cartea2022decentralised} & \checkmark & $\times$ & \checkmark & $\times$\\

    \bottomrule
    \end{tabular}
\end{table}

\begin{figure}[!t]
    \centering
    \includegraphics{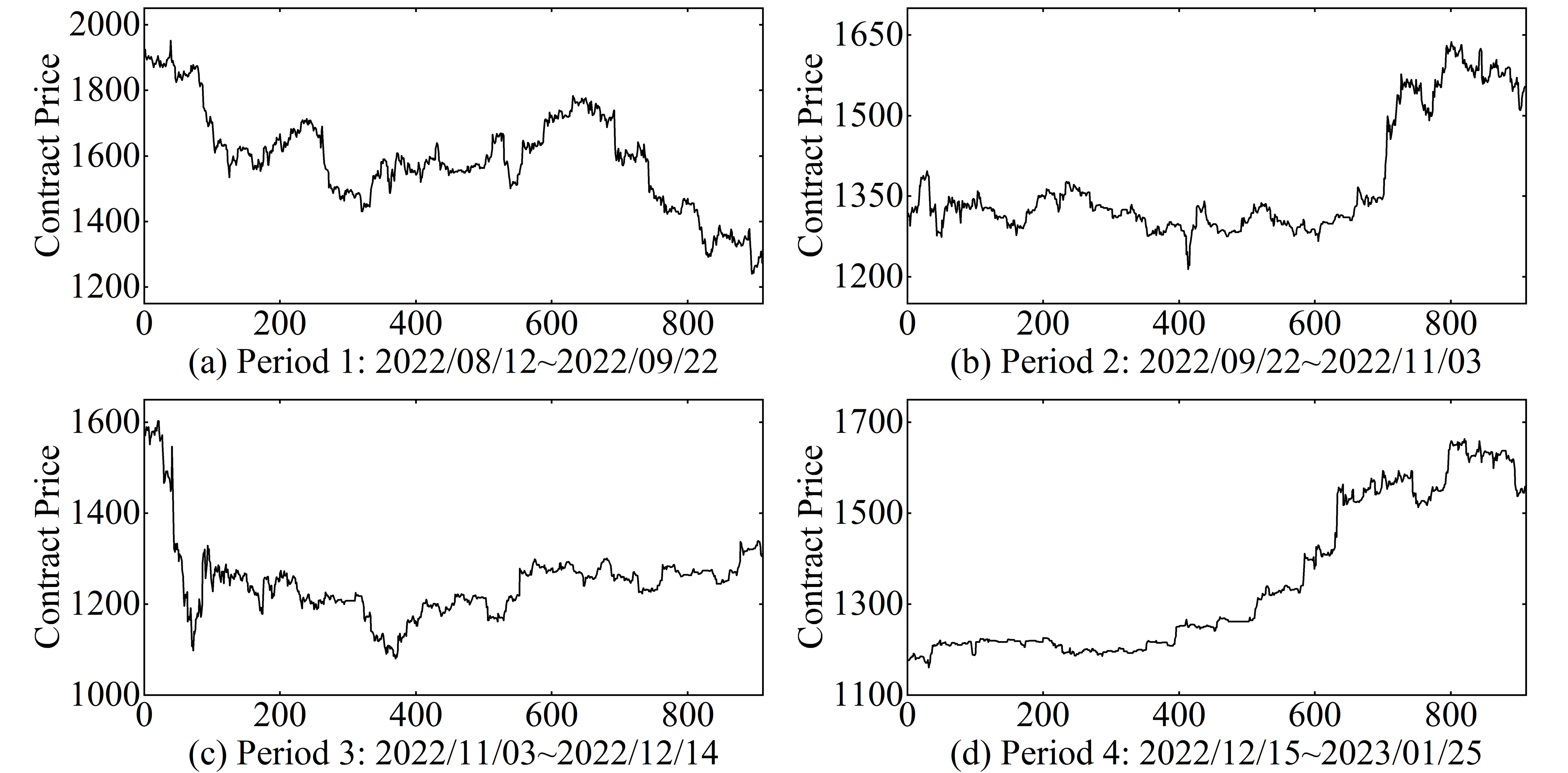}
    \caption{Contract price of ETH/USDC-0.3\% pool in different periods}
    \label{fig:4_period_price}
\end{figure}

M1 is trained and tested on data from the ETH/USDC-0.3\% pool and ETH/USDT-0.3\% pool during the period from August 2, 2021, to January 25, 2023. The original dataset is obtained from the Subgraph (https://thegraph.com/hosted-service/subgraph/uniswap/uniswap-V3) and is divided into four periods, as detailed in Table~\ref{tab:dataset}. Each period comprises 8000 hours of data for training, 1000 hours for validation, and 1000 hours for testing. Figure~\ref{fig:4_period_price} displays the contract price in the four test periods for the ETH/USDC-0.3\% pool. Due to the similar general trend of the contract price in the ETH/USDT-0.3\% pool, it is not shown here. For the hyperparameters in M2$\sim$M4, we select the best hyperparameters based on their performance on the test set. It is important to note that this approach yields stronger baselines than those encountered in real-world scenarios since the data in the test set is unknown in practice. While the hyperparameters in M1 are not fine-tuned, early stopping is employed to prevent overtraining. Detailed information about important hyperparameters and further implementation details for all methods can be found in Appendix~\ref{set:hyper}. For a more comprehensive understanding of the implementations of the aforementioned methods, please refer to our code on GitHub: https://github.com/HaochenZhang717/Uniswap-v3.git. The repository will become publicly accessible upon the acceptance of this paper.

\begin{table}[!t]
\centering
  \caption{Partition of Dataset}
  \label{tab:dataset}
  \begin{tabular}{cccc}
    \hline
            & training set            &   validation set          & test set\\
    \hline
    (1) & 2021/08/02$\sim$2022/07/01  &   2022/07/01$\sim$2022/08/11  &  2022/08/12$\sim$2022/09/22\\
    (2) & 2021/09/12$\sim$2022/08/11  &   2022/08/12$\sim$2022/09/22  &  2022/09/22$\sim$2022/11/03\\
    (3) & 2021/10/24$\sim$2022/09/22  &   2022/09/22$\sim$2022/11/03  &  2022/11/03$\sim$2022/12/14\\
    (4) & 2021/12/05$\sim$2022/11/03  &   2022/11/03$\sim$2022/12/14  &  2022/12/15$\sim$2023/01/25\\
  \hline
\end{tabular}
\end{table}

To evaluate the performance of different method, relative PnL is adopted herein. relative PnL in a period is defined as
\begin{displaymath}
    \mathrm{PnL}_{relative} = \frac{\sum_{t=0}^{T-1} r_t}{l_0},
\end{displaymath}
where $l_0$ denotes the initial fund, and the definition of $r_t$ can be seen in Section~\ref{reward_define}, which can be different depending hedging is used or not.


\begin{figure}[H]
  \centering
  \includegraphics{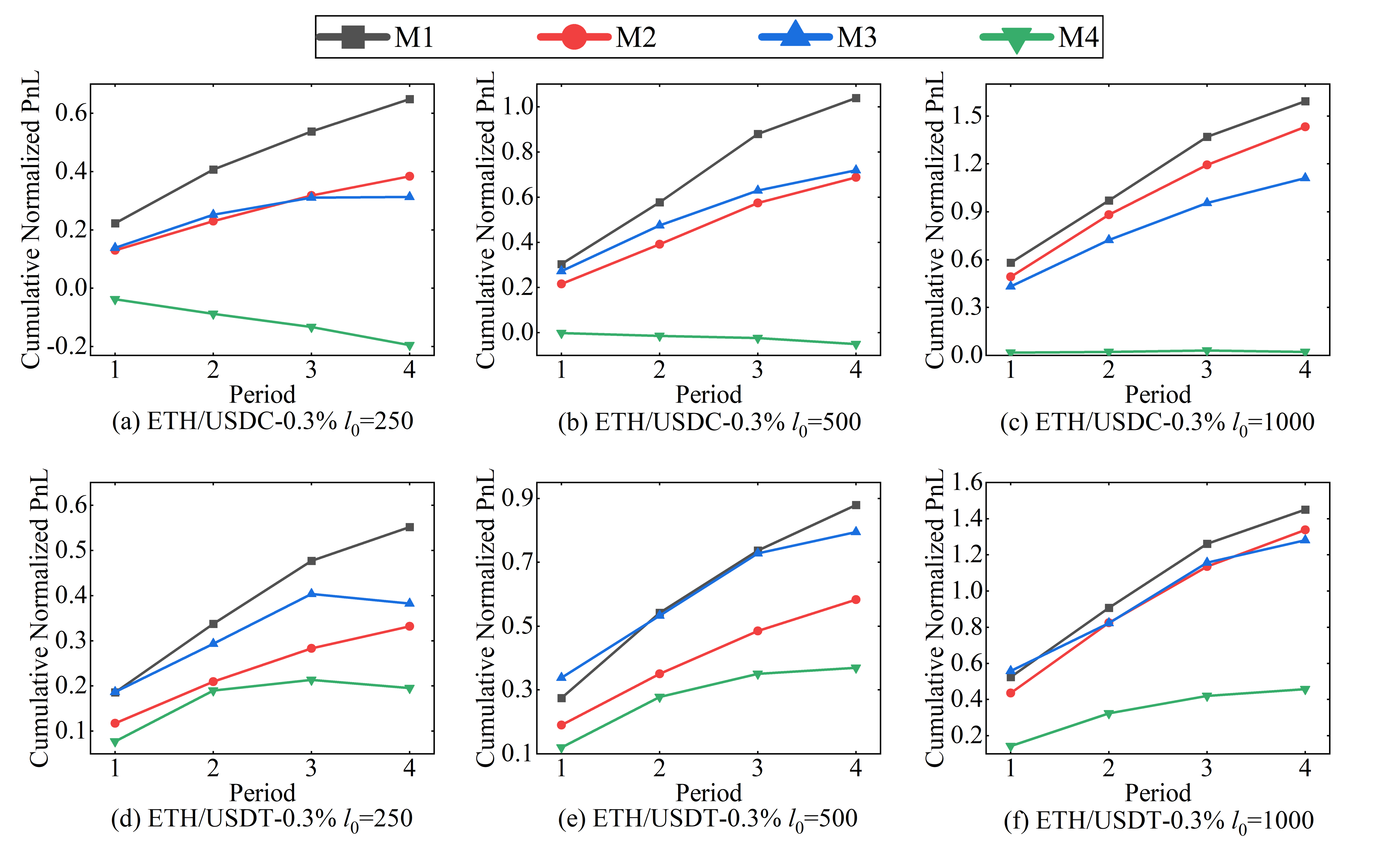}
  \caption{Cumulative normalized PnL of different methods from Period 1 to Period 4 in ETH/USDT 0.3\% pool and ETH/USDC 0.3\% pool with different initial fund $l_0=250,500,1000$}
  \label{fig:efficacy}
\end{figure}

\begin{figure}[!t]
    \centering
    \includegraphics{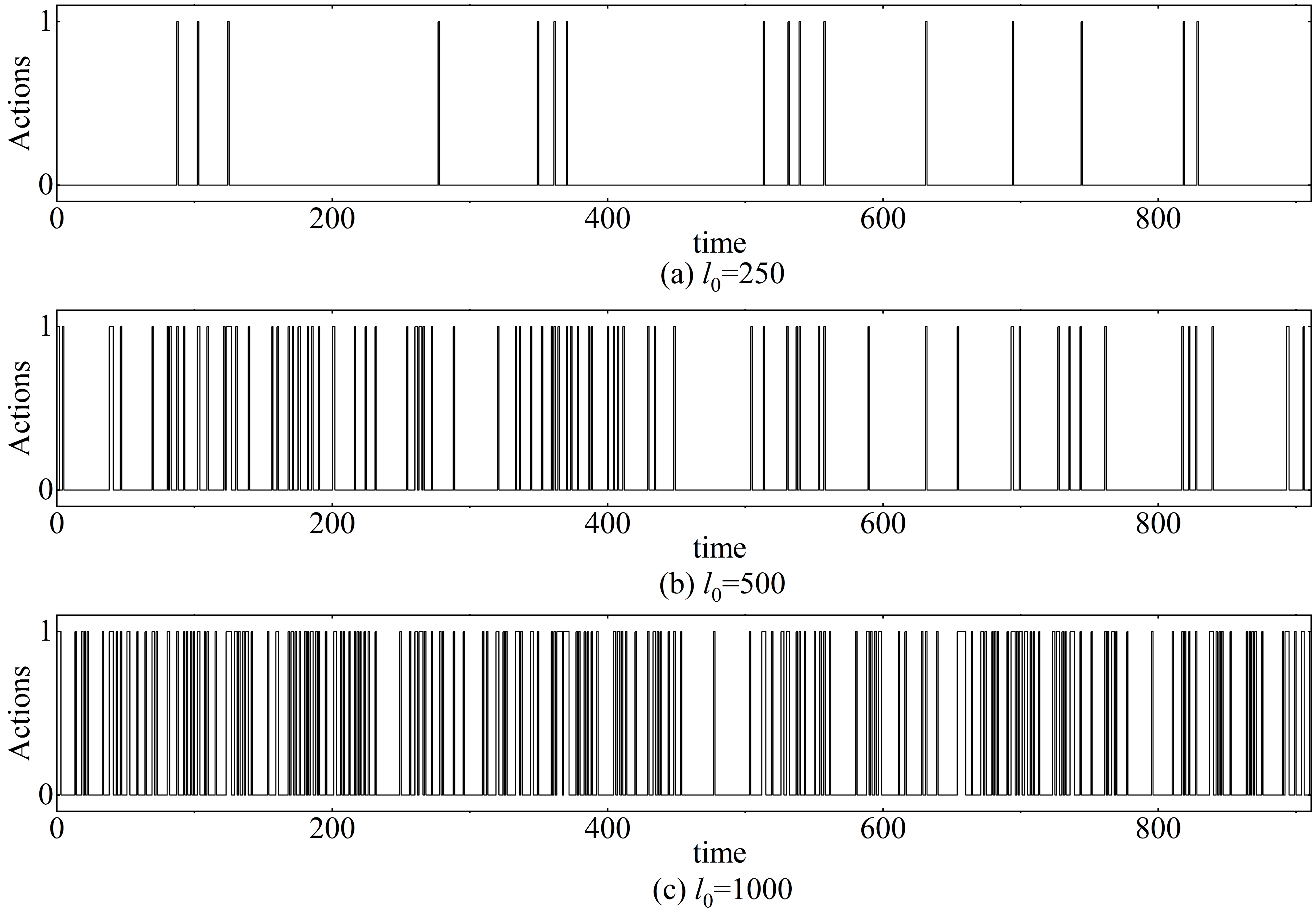}
    \caption{Actions made by M1 in Period 1 in ETH/USDT-0.3\% pool (M1 also optimizes liquidity spread, but the result shows it always prefers narrow interval with width equals 1)}
    \label{fig:action_period1}
\end{figure}

\subsection{Result and Discussion}

This section's first part discusses the efficacy of the proposed method from two perspectives. Firstly, it is shown that M1 has a superiority over M2$\sim$M4 in the sense that M1 is more profitable when hedging is combined with liquidity provision. Secondly, it is demonstrated that the RL agent can hardly learn a good strategy without hedging, indicating that the combination of hedging and liquidity provision can reduce the learning difficulty for the RL agent. The second part of this section shows the optimal LP behavior with different initial funds. As strictly optimal strategies are challenging to determine, M1's strategy is taken as an approximation.

When market risk is hedged by shorting the rebalancing portfolio, M1 is the most rewarding strategy, as depicted in Figure~\ref{fig:efficacy}. M1$\sim$M4 are tested with varying initial funds. It is observed that the cumulative relative PnL of M1 increases approximately from 0.6 to 1.5 as $l_0$ increases from 250 to 1000 through 500, indicating that investments with more initial funds are more profitable. As $l_0$ increases, the gap between M1 and the baselines becomes smaller. Note that a significant difference between M1 and the baselines is that M1 can adaptively control the hold period, whereas baselines cannot. The smaller gap between the relative PnL obtained by M1 and those obtained by baselines with larger $l_0$ indicates that it is less important for LPs to possess the knowledge of the optimal time to reallocate the liquidity position when LPs have a large mount of capital and gas fees are relatively cheap. This is a natural outcome since an LP can frequently reallocate the liquidity position to ensure the contract price remains within the range of their liquidity interval after each swap when the LP has infinite initial funds, rendering gas fees negligible. 
Conversely, when an LP's initial funds are limited (i.e., when gas fees are relatively expensive), the performance gap between baselines and M1 becomes larger. In this case, deciding to adjust or hold the liquidity position becomes more crucial for LPs.

\begin{table}[!t]
\centering
  \caption{Relative PnL of M1 with hedging and without hedging}
  \label{tab:rl_2_cases}
  \begin{tabular}{cccc|cc}
    \hline
    & & \multicolumn{2}{c|}{ETH/USDC-0.3\%} & \multicolumn{2}{c}{ETH/USDT-0.3\%}\\
    & & M1 & M1$\dag$ & M1 & M1$\dag$\\
    \hline
    \multirow{4}{1.5cm}{$l_0=250$} & period 1 & 0.223 & -0.259 & 0.186 & -0.259 \\
                                   & period 2 & 0.183 & 0.034 & 0.151 & 0.079 \\
                                   & period 3 & 0.130 & -0.131 & 0.139 & -0.141 \\
                                   & period 4 & 0.110 & 0.062 & 0.074 & 0.029 \\

    \hline
    \multirow{4}{1.5cm}{$l_0=500$} & period 1 & 0.305&-0.291&0.275&-0.270 \\
                                   & period 2 & 0.273&0.043&0.267&0.020 \\
                                   & period 3 & 0.302&-0.178&0.195&-0.125 \\
                                   & period 4 & 0.159&0.023&0.143&0.080 \\

    \hline
    \multirow{4}{1.5cm}{$l_0=1000$} & period 1 & 0.582 & -0.283&0.525&-0.278 \\
                                   & period 2 &0.388&0.061&0.383&0.050 \\
                                   & period 3 &0.400&-0.119&0.354&-0.105 \\
                                   & period 4 &0.222&0.079&0.189&0.071 \\
    \hline
\end{tabular}

\begin{tablenotes}
   \centering
    \footnotesize               
    \item[1] $\mathrm{M1}^\dag$ means Dueling DDQN is trained with the reward function defined \\ as PnL when hedging is not used as in section~\ref{reward_define}.          
\end{tablenotes}

\end{table}

\begin{table}[!t]
\centering
  \caption{Detailed test performance in terms of trading fee, gas fee, and LVR in ETH/USDC-0.3\% pool and ETH/USDT-0.3\% pool with $l_0=250$ (highest trading fee, lowest gas fee, lowest LVR, and highest PnL are in bold.)}
  \label{tab:details_in_both}
  \begin{tabular}{cccccc|cccc}
    \toprule
       &  & \multicolumn{4}{c|}{ETH/USDC-0.3\%}  &  \multicolumn{4}{c}{ETH/USDT-0.3\%}   \\
       &  & M1 &  M2  & M3  & M4 & M1 &  M2  & M3  & M4  \\
    \hline
         \multirow{5}{1.5cm}{Period 1}  &  relative trading fee & 0.691 & 0.63  & \textbf{0.774} & 0.088 & 0.605 & 0.614 & \textbf{0.921} &0.402\\
                                        &  relative gas fee     & \textbf{0.113}  & 0.207  & 0.293  & 0.120 &\textbf{0.096}&0.207&0.293&0.133\\
                                        &  relative LVR         & 0.205  & 0.205  & 0.248  & \textbf{0.030}&0.199&0.211&0.316&\textbf{0.139}\\
                                        &  relative PnL         & \textbf{0.373}&0.218&0.232&-0.062&\textbf{0.310}&0.196&0.311&0.130 \\

    \cline{2-10}
        \multirow{5}{1.5cm}{Period 2}  &  relative trading fee & 0.607 & 0.453 & \textbf{0.611} & 0.052&0.503&0.459&\textbf{0.587}&0.452\\
                                       &  relative gas fee     & \textbf{0.118} & 0.153 & 0.253 & 0.120&\textbf{0.089}&0.147&0.253&0.140\\
                                       &  relative LVR         & 0.182 & 0.132 & 0.168 & \textbf{0.015}&0.161&0.159&0.156&\textbf{0.125}\\
                                       & relative PnL&\textbf{0.307}&0.168&0.190&-0.083&\textbf{0.252}&0.154&0.178&0.187\\

    \cline{2-10}
          \multirow{5}{1.5cm}{Period 3}  & relative trading fee & 0.541 & \textbf{0.569} & 0.550 & 0.089&0.686&0.391&0.777&\textbf{0.295}\\
                                         & relative gas fee     & \textbf{0.104} & 0.167 & 0.280 & 0.120&0.119&\textbf{0.060}&0.280&0.153\\
                                         & relative LVR         & 0.220 & 0.257 & 0.174 & \textbf{0.045}&0.335&0.209&0.313&\textbf{0.103}\\
                                         & relative PnL&\textbf{0.217}&0.145&0.096&-0.076&\textbf{0.232}&0.122&0.184&0.039\\

    \cline{2-10}
        \multirow{5}{1.5cm}{Period 4}  &  relative trading fee & \textbf{0.370} & 0.318 & 0.365 & 0.031&\textbf{0.463}&0.348&0.373&0.149\\
                                       &  relative gas fee     & \textbf{0.064} & 0.100 & 0.253 & 0.120&\textbf{0.068}&0.107&0.293&0.120\\
                                       &  relative LVR         & 0.121 & 0.108 & 0.107 & \textbf{0.014}&0.271&0.159&0.115&\textbf{0.059}\\
                                       & relative PnL &\textbf{0.185}&0.109&0.004&-0.104&\textbf{0.124}&0.082&-0.035&-0.030\\ 

  \bottomrule
\end{tabular}
\end{table}

Table~\ref{tab:rl_2_cases} presents the relative PnL of M1 in two scenarios: when it is combined with hedging and when it is not. This result highlights a significant benefit of adopting hedging and training the Dueling DDQN agent with the corresponding reward function in the sense that the relative PnL is less influenced by the trend of the contract price. Notably, the contract price exhibits a general downward trend in Period 2 and Period 4, while showing an upward trend in Period 1 and Period 3, as depicted in Figure~\ref{fig:4_period_price}. This trend aligns with the relative PnL obtained by M1$\dag$, which is negative in Period 2 and Period 4, but positive in Period 1 and Period 3. However, the relative PnL gained by M1 is not correlated with the direction of the price movement, demonstrating the benefit derived from hedging.



Table~\ref{tab:details_in_both} displays the relative trading fee, relative gas fee, and relative LVR in both pools with $l_0=250$. These three relative terms are obtained by dividing the actual values by the initial fund $l_0$. Among the four methods tested herein, M1 learns the most profitable strategy. However, upon closer examination of the trading fee, gas fee, and LVR, it becomes evident that M1 does not consistently obtain the highest trading fee or incur the lowest gas fee and LVR. Instead, M1 finds the best trade-off between the reward of providing liquidity and its associated costs—namely, the sum of the gas fee and LVR.

An intriguing observation is that whenever M1 reallocates the liquidity position with initial funds of $l_0=250$, $500$, and $1000$, it consistently favors the narrowest price interval with the smallest width, as depicted in Figure~\ref{fig:action_period1}. In other words, despite the action space being defined as $\mathcal{A}=\{0,1,...N\}$, M1 consistently opts for either $a_t=0$ or $a_t=1$. Nevertheless, the difference in M1's behavior across different initial fund values lies in the frequency of liquidity position reallocation—it occurs less frequently with lower initial funds.

\section{Conclusion\label{sec:discussion and conclusion}}

This study introduces an adaptive uniform interval reallocation strategy for providing liquidity in Uniswap V3 by integrating hedging and deep reinforcement learning. The counterbalancing effect of the rebalancing strategy can mitigate risks for LPs in Uniswap V3, though leaving behind a predictable residue in terms of LVR. Deep reinforcement learning is employed to address two practical challenges for LPs: determining the optimal timing to adjust the liquidity position and establishing the appropriate width for the liquidity interval. Simulations based on historical data from the ETH/USDC 0.3\% pool and ETH/USDT 0.3\% pool empirically validate the viability of applying reinforcement learning methods to optimize liquidity provision in Uniswap V3. The Dueling DDQN model yields profits that are 9\% to 69\% higher than those of the strongest baseline across different pools with different initial funds.

\bibliographystyle{unsrt}  
\bibliography{references}

\appendix

\section{Implementation Details of The Proposed Method and Baselines \label{set:hyper}}

\begin{algorithm}[H]
\caption{Exponential Weights Adaptive Strategy (EWA)}\label{alg:bandit}
\begin{algorithmic}[1]
\STATE {Input $N\in \mathbb{N_{+}},\ \eta> 0,\ T_{re} \in \mathbb{N_{+}}$}
\STATE {Initialize $r_{0}(n)\gets 0 \ \mathrm{for \ all \ } n\in \left[N\right]$}
\FOR{$1\le t\le T$}
\IF{$\mathrm{mod}\left(t, T_{re}\right)==0$}
\STATE $p_{t}\left( n\right)\gets \frac{\mathrm{exp}\left(\eta \sum_{0\le\tau\le t}r_{\tau}\left(n\right) \right)}{\sum _{1\le\mu\le N}{\mathrm{exp}\left(\eta\sum_{0\le\tau\le t}r_{\tau}\left(\mu\right) \right)}} $
\STATE {invest $p_t(n)$ proportion of funds into a liquidity position with width n, and get reward $r_t(n)$}
\ENDIF
\ENDFOR
\end{algorithmic}
\label{alg1}
\end{algorithm}

Hyperparameters of EWA are choosen by maximizing its performance on validation data. Hyperparameters of uniform $\tau$-reset strategy are determined by using testing set. In reality, testing data is unkonwn, so this setting gives uniform $\tau$-reset strategy an advantage, thus this baseline is stronger than in practice. Hyperparamters of Dueling DDQN are provided in Table~\ref{tab:hp_dqn}, respectively. Hyperparameters of Dueling DDQN are fixed for all four periods. In this paper's experiment, hypereparameters of Dueling DDQN are tuned roughly, it is believed that further fine tuning is likely to improve their performance. 

\begin{table}[H]
\caption{hyperparameters of EWA}
    \centering
    \begin{tabular}{cccccccccccc}
    \toprule
    & & \multicolumn{3}{c}{$l_0=250$} &  \multicolumn{3}{c}{$l_0=500$}  &  \multicolumn{3}{c}{$l_0=1000$}  \\
    & & $N$ & $\eta$ & $T_{re}$ & $N$ & $\eta$ & $T_{re}$& $N$ & $\eta$ & $T_{re}$ \\
    \midrule
    
    \multirow{4}{2.7cm}{ETH-USDC 0.3\%} & Period 1 &10&1&21& 10&1&14 & 10 & 1 & 6\\
                                        & Period 2 &10&10&24& 10&10&24&10 & 10 & 9 \\
                                        & Period 3 &10&1&22& 10&4&15&10 & 1 & 13 \\
                                        & Period 4 &10&7&24& 10 & 1 & 21&10&1&18 \\
    \cline{2-11}
    \multirow{4}{2.7cm}{ETH-USDC 0.3\%} & Period 1 & 10&1&21&10&1&6&10&1&6 \\
                                        & Period 2 & 10&10&24&10&10&24&10&10&12 \\
                                        & Period 3 & 10&1&22&10&7&22&10&10&3 \\
                                        & Period 4 & 10&7&21&10&1&21&10&1&21 \\
    \bottomrule
    \end{tabular}
    \label{tab:hp_ewa}
\end{table}

\begin{table}[H]
\caption{hyperparameters of uniform $\tau$-reset strategy}
    \centering
    \begin{tabular}{ccccc}
    \toprule
    & & $l_0=250$ & $l_0=500$ & $l_0=1000$ \\
    & & $\tau$ & $\tau$ & $\tau$ \\
    \midrule
    
    \multirow{4}{2.2cm}{ETH-USDC 0.3\%} & Period 1 &6&4& 1  \\
                                        & Period 2 &5&2& 1  \\
                                        & Period 3 &6&3& 2  \\
                                        & Period 4 &4&3& 1  \\
    \cline{2-5}
    \multirow{4}{2.2cm}{ETH-USDT 0.3\%} & Period 1 & 6&4&1 \\
                                        & Period 2 & 5&2&1 \\
                                        & Period 3 & 10&3&1 \\
                                        & Period 4 & 4&3&1 \\
    \bottomrule
    \end{tabular}
    \label{tab:hp_tau}
\end{table}

\begin{table}[H]
\caption{hyperparameters of Dueling DDQN}
    \centering
    \begin{tabular}{ccc}
    \toprule
    \textbf{hyperemeters} & \textbf{values} \\
    \midrule
    hidden unites & $[64,64]$ \\
    activation & ReLU \\
    final activation & None \\
    learning rate & $10^{-4}$ \\
    batch size & 256 \\
    buffer size & $10^{6}$ \\
    discounted factor & 0.9 \\
    target update rate & 0.01 \\
    gradient clipping norm & 0.7 \\   
    \bottomrule
    \end{tabular}
    \label{tab:hp_dqn}
\end{table}

\end{document}